\documentclass{article}

\PassOptionsToPackage{numbers, compress}{natbib}

\usepackage[preprint]{neurips_2024}

\usepackage{graphicx}
\usepackage[utf8]{inputenc} 
\usepackage[T1]{fontenc}    
\usepackage{hyperref}       
\usepackage{url}            
\usepackage{booktabs}       
\usepackage{amsfonts}       
\usepackage{nicefrac}       
\usepackage{microtype}      
\usepackage{xcolor}         
\usepackage{mathtools}

\usepackage{amsfonts,bm,physics}

\usepackage{amsmath}
\usepackage{multirow}
\usepackage{cleveref}

\usepackage{color, colortbl}
\definecolor{Gray}{gray}{0.9}
\definecolor{lightgray}{gray}{0.5}
\definecolor{verylightgray}{gray}{0.7}
\definecolor{veryverylightgray}{gray}{0.9}

\definecolor{darkgreen}{rgb}{0, 0.4, 0}
\definecolor{darkred}{rgb}{0.7, 0, 0}
\definecolor{darkblue}{rgb}{0.0, 0.0, 0.7}

\newcommand{\pstud}{q_s}
\newcommand{\pteach}{q_t}

\newcommand{\temiloss}{\mathcal{L}_\text{\sc{temi}}}
\newcommand{\scanloss}{\mathcal{L}_\text{\sc{scan}}}

\title{Scaling Up Deep Clustering Methods \\ Beyond ImageNet-1K}

\author{
Nikolas Adaloglou \\
  Heinrich Heine University of Dusseldorf\\
    \texttt{nikolaos.adaloglou@hhu.de} \\
 \And  Felix Michels \\
  Heinrich Heine University of Dusseldorf \\
  \texttt{felix.michels@hhu.de} \\
 \And  Kaspar Senft\thanks{Authors contributed equally.} \\
  Heinrich Heine University of Dusseldorf\\
  \texttt{kaspar.senft@hhu.de} \\
  \And  Diana Petrusheva\footnotemark[1] \\
  Heinrich Heine University of Dusseldorf\\
  \texttt{diana.petrusheva@hhu.de} \\ 
  \And  Markus Kollmann \\
  Heinrich Heine University of Dusseldorf\\
  \texttt{markus.kollmann@hhu.de} \\   }

\begin{document}

\maketitle
\begin{abstract}
Deep image clustering methods are typically evaluated on small-scale balanced classification datasets while feature-based $k$-means has been applied on proprietary billion-scale datasets. In this work, we explore the performance of feature-based deep clustering approaches on large-scale benchmarks whilst disentangling the impact of the following data-related factors: i) class imbalance, ii) class granularity, iii) easy-to-recognize classes, and iv) the ability to capture multiple classes. Consequently, we develop multiple new benchmarks based on ImageNet21K. Our experimental analysis reveals that feature-based $k$-means is often unfairly evaluated on balanced datasets. However, deep clustering methods outperform $k$-means across most large-scale benchmarks. Interestingly, $k$-means underperforms on easy-to-classify benchmarks by large margins. The performance gap, however, diminishes on the highest data regimes such as ImageNet21K. Finally, we find that non-primary cluster predictions capture meaningful classes (i.e. coarser classes).
\end{abstract}

\section{Introduction}
For over a decade, the ImageNet-1K benchmark \cite{deng2009imagenet,russakovsky2015imagenet} has served as the core downstream dataset in a plethora of computer vision tasks. Examples include image classification \cite{resnet}, representation learning \cite{simclr}, semi-supervised learning \cite{sohn2020fixmatch}, and recently image clustering \cite{sela}. Image clustering, also known as unsupervised image classification, refers to algorithmically grouping visual stimuli into discrete concepts called clusters \cite{battleday2020capturing,adaloglou2024rethinking}. Grouping images without human supervision has many applications, including unsupervised out-of-distribution detection \cite{sehwag2021ssd}, image generation \cite{bao2022why}, and large-scale dataset pruning \cite{dinov2,abbas2024concept_clusters}. 


The advent of deep neural networks has led to the emergence of deep image clustering \cite{chang2017deep} and feature-based clustering. Examples of feature extractor pretraining include visual self-supervision \cite{simclr,dino} or natural language supervision \cite{radford2021clip}. The most widely established feature-based clustering approach to date is $k$-means \cite{lloyd1982kmeans} due to its scalable nature. Feature-based $k$-means has been successfully employed to billion-scale proprietary vision datasets \cite{dinov2}. Concurrently, deep clustering methods such as TEMI \cite{temi} have achieved state-of-the-art clustering performance on ImageNet-1K \cite{deng2009imagenet,russakovsky2015imagenet}, outperforming $k$-means by significant margins \cite{temi}. This discrepancy in application scales raises questions about the applicability of deep clustering methods beyond the scope of balanced classification datasets (ImageNet-1K) as well as their comparability to $k$-means in real-world, large-scale scenarios. To address this, we start off by first presenting the key inherent and non-inherent limitations regarding the task of image clustering.

\textbf{Inherent limitations.} The first inherent limitation is that the valid computation of the clustering accuracy requires setting the number of clusters to the number of human-derived classes, the \textit{ground truth (GT)}. Second, similar to image classification, real-world images frequently contain multiple objects of interest, which is not captured when considering only the top 1 class prediction. Third, biases in the data annotation procedure can lead to systematic inaccuracies \cite{beyer2020done_imagenet}, even at small data regimes \cite{barz2020purging_cifar}. Human annotators may assign a particular label that seems plausible when presented in isolation or create visually indistinguishable groups based on arbitrary class distinctions (i.e. the class ``laptop computer'' and ``notebook computer'' coexist in ImageNet-1K) \cite{beyer2020done_imagenet, battleday2020capturing}. Finally, unlike supervised approaches, clustering algorithms cannot accurately capture highly coarse labels solely based on visual input, such as carnivoran mammals \cite{adaloglou2024rethinking,cifar}.

\textbf{Non-inherent limitations.} More importantly, current clustering benchmarks have specific constraints that are non-inherent limitations of the task at hand. First, the majority of methods are being developed on small-scale datasets such as CIFAR10 and CIFAR100 \cite{cifar}, with only a limited number of methods reporting results on ImageNet-1K \cite{scan,sscn,temi}. It is thus unclear whether these methods apply to large-scale benchmarks, often considering their non-scalable nature. Spectral clustering with an $O(N^3)$ time complexity \cite{ng2001spectral} is one example, where $N$ is the number of samples. Second, existing benchmarks are primarily balanced classification datasets, wherein all samples are uniformly distributed among the GT labels. While deep image clustering methods leverage this information using regularization techniques, classical machine learning methods (i.e. $k$-means) do not have this flexibility and tend to create imbalanced clusters \cite{scan}. New challenging benchmarks that mimic real-world scenarios would aid in bridging the gap between ImageNet-1K and billion-scale proprietary datasets. 

In this paper, we perform a comprehensive experimental study on large-scale clustering methods and benchmarks. We create various new clustering benchmarks based on ImageNet21K, focusing on the effect of individual factors, namely class imbalance, class granularity, easily ``classifiable'' classes, and the ability to capture multiple labels. We show that evaluating only on balanced classification datasets treats $k$-means unfairly, yet deep clustering methods still outperform $k$-means across most benchmarks besides ImageNet-1K. The gap, however, diminishes on large-scale benchmarks with more than 7K classes. We find that i) $k$-means attains inferior performance on easy-to-classify benchmarks by large margins, and ii) non-primary cluster predictions of clustering methods capture meaningful classes such as coarser or coexisting labels. 

\section{Related work}

\textbf{Large-scale image clustering methods.} Deep image clustering consists of learning the label-related representations and the cluster assignments \cite{temi}, either simultaneously (single-stage methods) \cite{sela,sscn} or sequentially (multi-stage methods) \cite{scan}. We consider a method to be large-scale if clustering results are reported on Imagenet-1K \cite{deng2009imagenet}. In SeLa \cite{sela}, the authors design a two-step framework, which alternates between estimating the pseudo-label assignment matrix and representation learning. In \cite{sscn}, the authors present a single-stage end-to-end method that employs a variant of the cross-entropy loss. However, such single-stage methods require dataset-dependent hyperparameter tuning and are computationally expensive to iterate on large scales.

Contrastive learning \cite{simclr} has been a major breakthrough for image clustering. SCAN \cite{scan} was the first large-scale method to isolate representation learning from learning the cluster assignments. By first employing contrastive learning on the training data, \citet{scan} show that nearest neighbors (NNs) in feature space likely share the same label. Based on the same principle, \citet{temi} have recently shown that a) features from multiple pre-trained feature extractors (i.e. DINO \cite{dino}, MSN \cite{msn}) can be used and b) learning the cluster assignments does not require image augmentations as opposed to SCAN \cite{scan}. This enables pre-computing the image features similar to feature-based $k$-means. 

Recent clustering methods primarily focus on learning informative feature representation by sequentially combining masked image modeling with contrastive learning \cite{alkin2024mae_refine, singh2023mae_prepretraining}, which has been recently coined as pre-pretraining \cite{singh2023mae_prepretraining}. As we focus on learning the cluster assignments, our work is closer to TEMI \cite{temi}, an alternative to feature-based $k$-means. Nevertheless, deep learning-based clustering methods have only been applied up to the scale of ImageNet-1K ($\approx 1.2M$ images). Using recent advancements in GPU-accelerated similarity search and a distributed implementation of $k$-means \cite{johnson2019billion_faiss,douze2024faiss}, \citet{dinov2} scaled up feature-based $k$-means \cite{lloyd1982kmeans} to 100K clusters and 1.2B images. The divergence in the above-mentioned data regimes makes it unclear whether other methods can be scaled beyond ImageNet-1K, which we aim to address in this work using Imagenet21K.

\textbf{Leveraging ImageNet21K class structure.} The set of ImageNet21K classes and its semantic tree has been used in a plethora of downstream applications, such as hierarchical image classification \cite{novack2023chils}, object detection \cite{zhou2022detecting20k} and zero-shot learning \cite{xian2018zero,chen2021hsva,yi2022hierarchical}. \citet{zhou2022detecting20k} apply CLIP \cite{radford2021clip} on the predicted bounding boxes to generate diverse labels based on ImageNet21K for object detection. To increase the generalization capability to unseen classes, \citet{yi2022hierarchical} exploit the semantic structure of ImageNet21K during vision-language training for zero-shot learning. \citet{novack2023chils} demonstrated the superior performance of CLIP (zero-shot classification) leveraging more fine-grained classes than the ground truth. The discussed works have explored the ImageNet21K semantic structure to boost downstream task performance. Instead, we leverage the ImageNet21K semantic tree to vary the ground truth class granularity.

\section{Background, materials and methods}
\textbf{Existing benchmarks and pre-trained models.} The most challenging large-scale clustering benchmark to date is ImageNet-1K \cite{deng2009imagenet,russakovsky2015imagenet}, which consists of $C$=1000 balanced classes ($p(c)=\mathcal{U}(\{1, .., C \})$). To the best of our knowledge, the only class imbalance that has been proposed is halving the number of samples from the odd class indices \cite{ding2023mlc}. In contrast to \cite{ding2023mlc}, a) we do not modify the validation set as we want to quantify the performance degradation attributed to the imbalance, and b) we apply it on ImageNet-1K instead of CIFAR10 and CIFAR100 \cite{cifar}. We refer to this benchmark as \textit{ImageNet-1K ODD}. All the newly-created benchmarks are based on the preprocessed version of ImageNet21K (winter version 2021) as in \cite{ridnik2021imagenet}. This version of ImageNet21K has $\approx11.06$M samples that are non-uniformly distributed across 10450 non-mutually exclusive classes. These classes can be organized in a semantic tree based on the WordNet hierarchy \cite{miller1995wordnet,xue2011nltk}. To imitate a real-life scenario where GT labels and additional data are scarce, we assume no access to external data or models trained on external data. Unless otherwise specified, we use the iBOT ViT-L \cite{zhou2021ibot} pre-trained on ImageNet21K and MAE Refined ViT-H \cite{alkin2024mae_refine} (\textit{MAE-R}) on ImageNet-1K (current state-of-the-art).

\textbf{Feature-based image clustering methods.} Apart from distributed $k$-means \cite{lloyd1982kmeans,douze2024faiss}, we identify and use two scalable state-of-the-art methods based on the framework of \cite{temi}, namely TEMI and SCANv2 \cite{temi}, which we describe below. TEMI first mines the nearest neighbors (NN) in feature space for each sample $x$ in the dataset $D$. During training, TEMI randomly samples $x$ from $D$ and $x'$ from the set of NN of $x$. TEMI employs a teacher and student head $h_t(\cdot)$ and $h_s(\cdot)$ that have an identical architecture but differ w.r.t.\ their parameters as the teacher is only updated using an exponential moving average of $h_s(\cdot)$. Given a feature extractor $g(\cdot)$, the student and teacher consist of multiple independent heads $h^{i}_s(z)$, $h^{i}_t(z')$, where $i \in \{ 1, \dots, H \}$ is the head index and $z=g(x)$, $z'=g(x')$. Class probabilities are computed using a softmax function $\pstud(c|x)$ and $\pteach(c|x')$ and the following loss is minimized  
\begin{equation}
     \temiloss^i(x,x') \coloneqq -
     \underbrace{\frac{1}{H} \sum_{j=1}^H \sum_{c'=1}^C \pteach^j(c'|x)\pteach^j(c'|x')}_{\text{instance weighting}}
     \log \sum_{c=1}^C \frac{
    \left( \pstud^i(c|x) \pteach^i(c|x')\right)^{\beta}}
    {\tilde q_t^i(c)} \quad .
\label{eq:temi-loss}
\end{equation} 

In \Cref{eq:temi-loss}, $\tilde\pteach^i(c)$ is an estimate of the teacher cluster distribution $\mathbb{E}_{x \sim p_\text{data}}\left[\pteach^i(c|x)\right]$, which can be computed by an exponential moving average over mini-batches $B$ defined as $\tilde \pteach^i(c) \leftarrow \lambda \,\tilde \pteach(c) + (1 - \lambda) \frac{1}{\abs{B}}\sum_{x \in B} \pteach^i(c|x)$ for $\lambda \in (0,1)$. The scalar $\beta\in (0.5,1]$ avoids assigning all sample pairs to a single cluster (mode collapse) and is typically set to $0.6$. The training loss is symmetrized and averaged across heads. After training, the head with the lowest training loss is kept for evaluation. SCANv2 uses the same self-distillation framework but uses the semantic clustering loss as in \citet{scan}

\begin{equation}
    \scanloss^i(x,x') \coloneqq -
    \log \sum_{c=1}^C \pteach(c|x)\pstud(c|x') - \alpha 
    \underbrace{\sum_{c=1}^C \hat\pstud(c) \log\hat\pstud(c)}_{\text{Negative entropy of} \hat\pstud^i(c)},
    \label{eq:scan-loss}
\end{equation}
where $\alpha>0$ is a scalar and $\hat\pstud^i(c) = \frac{1}{\abs{B}}\sum_{x\in B} \pstud^i(c|x)$ is the current mini-batch estimate of the student cluster distribution. Unlike the TEMI objective, $\scanloss^i(x,x')$ only depends on the head number $i$. The entropy term of \Cref{eq:scan-loss} is upweighted by $\alpha=5$ and acts conceptually similar to $\tilde\pteach^i(c)$ in \Cref{eq:temi-loss}.

\section{New clustering benchmarks based on ImageNet21K}



\subsection{Quantifying the sensitivity to class imbalance}
Halving the number of odd class indices can only be meaningfully applied to balanced classification datasets. Instead, we create subsets based on the class histogram (number of samples per class). To create varying degrees of imbalance, we take multiple class percentiles $s$ around the median class frequency $m_s =[50-s, 50+s]$, where $s \in \{5,15,25,35,45\}$. As $s$ increases, the number of samples $N$ and classes $C$ also increase. Additionally, we create a highly imbalanced subset by taking a percentile of 10\% around the median ($m_{10}$), adding the 10\% most frequent classes, and comparing it against taking the same number of classes centered around the median. We call these frequency-based ImageNet21K benchmarks \textit{Imbalanced-2K} and \textit{Centered-2K}, respectively.

\subsection{Quantifying the sensitivity to the class granularity}

\textbf{Coarse benchmarks.} Class labels can be organized in a hierarchical semantic tree \cite{subero2020tree}, such as WordNet \cite{miller1995wordnet}. Based on WordNet, \citet{ridnik2021imagenet} have identified 11 hierarchy levels on ImageNet21K. A hierarchy of 1 corresponds to the most coarse level and 11 to the most fine-grained. Using a semantic tree, one can seamlessly find the linguistic hypernyms (semantic ancestors) for each GT class, such as going from the GT class ``chair'' to ``furniture''. By leveraging the ImageNet21K tree, we recursively map the GT classes to their semantic ancestor such that the maximum hierarchy depth is restricted to $d \in [1,.., 9]$, where the majority of the samples lie. Note that we do not exclude any samples at this step and leave classes with a lower hierarchy than $d$ intact. During the learning process of the clustering algorithm, we create as many groups as the number of coarsened labels and evaluate on the validation set with the coarse labels of maximum depth $d$.

\textbf{Fine-grained benchmarks.} Refining ground truth (GT) labels to more fine-grained classes is challenging compared to coarsening them. The simplest way to achieve maximum class granularity is to consider only images from classes without linguistic hyponyms (semantic descendants) in the semantic tree, known as leaf classes \cite{yang2024imagenetood}. We refer to these subsets as \textit{WordNet leaf} and \textit{ImageNet21K leaf}, based on their respective semantic trees. Unlike ImageNet21K, these subsets are mutually exclusive. To avoid excluding samples from non-leaf classes, we first use CLIP in tandem with the semantic tree to reannotate non-leaf samples. To this end, we perform zero-shot classification on the hyponyms to obtain a new refined label, such as furniture$\rightarrow$\{chair, desk, shelf, couch\}. This procedure is recursively applied, always selecting labels with the highest image-text similarity. We stop once a leaf class is assigned. We refer to this method as leaf hierarchical zero-shot label refining (\textit{leaf HZR}).


\begin{figure}
\begin{center}
  \includegraphics[width=1\columnwidth]{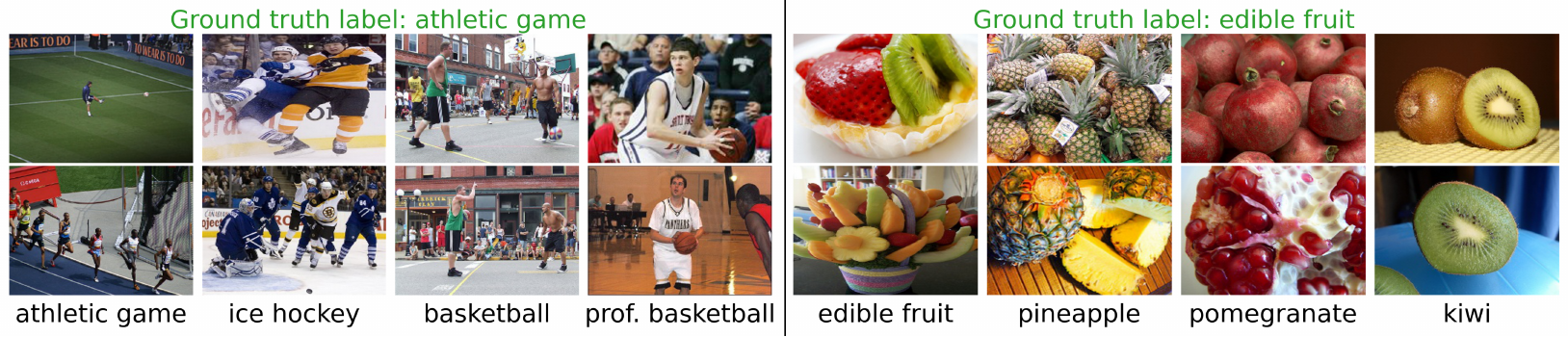}
    \caption{Reassessed ImageNet21K samples using parent hierarchical zero-shot label refining
(p-HZR). The original GT label is shown on top, while each column shows the reassessed label using openCLIP ViT-G \cite{openclip} in conjunction with the semantic tree.} 
    \label{fig:hzr}
\end{center}
\end{figure}
It is not, however, guaranteed that an image can always be meaningfully relabeled as one of its hyponyms. Analogously to a human annotator that can decide to keep the current label \cite{guo2017calibration, gawlikowski2023survey_uncertainty}, we add the parent class during zero-shot classification. In the previous example, we perform zero-shot classification using the following class set \{chair, desk, shelf, couch, furniture\}. By including the parent class at each stage of the recursive relabelling process, we allow CLIP to assign the same label to the image, partially mitigating the aforementioned issue, as illustrated in \Cref{fig:hzr}. We refer to this setup as parent HZR \textit{(p-HZR)}.


\subsection{Easy-to-classify benchmarks: model-based ImageNet21K subsets}
Here, we assess whether easy-to-classify samples are easy to cluster, as measured by high linear class separability \cite{wang2020understanding}. The intuition behind this is that if an image is not classified correctly, it is unlikely to be clustered successfully. We choose classes that are recognized with high ACC from pre-trained models with additional information during pre-training. To realize this, we use a subset of the top-$1K$ class accuracies from three pre-trained models: 1) a vision-language model (OpenCLIP ViT-G \cite{openclip}), 2) a visual self-supervised model (DINOv2 ViT-g \cite{dinov2}), and 3) an Imagenet21K-supervised model (ViT-L \cite{rw2019timm}). More precisely, we measure the top-1K accuracies per class on the validation set after linear probing. We name the created subsets as \textit{CLIP-1K}, \textit{DINOv2-1K}, \textit{Sup-1K}. We compare the model-based subsets with random subsets of 1K classes, ImageNet-1K, and their class union. 


\subsection{Multi-label clustering benchmarks and metrics} 
Here, we investigate whether clustering methods capture additional concepts, which is particularly useful for non-mutually exclusive datasets and large-scale datasets with multiple objects of interest \cite{beyer2020done_imagenet}. For each image, we consider a set of labels $S_L$ such that $S_L=\{ s_1, .. , s_{L} \}$, where $ S_L \subset S_C$ (set of GT classes). For hierarchically structured datasets, we set $S_L$ to be all semantic ancestors of the GT label. For example, an image of a guitarist (GT) can be classified as a musician or person, something that is not captured by existing clustering metrics. We aim to measure this with the top 1$\rightarrow$L ACC similar to \citet{beyer2020done_imagenet}.

In practice, we adopt ImageNet21K \cite{ridnik2021imagenet} and the reassessed multi-label validation set of ImageNet-1K called ``ReaL'' \cite{beyer2020done_imagenet} as benchmarks. We also assess the quality of the top 5 predictions as in \cite{scan}. To estimate the Hungarian one-to-one mapping $f$ \cite{kuhn1955hungarian}, we use the top 1 prediction, and select a single label from $S_L$. After computing $f$, the top 5 predictions are mapped to GT classes, and the top $5\rightarrow$1 and top $5\rightarrow L$ accuracies are computed. For ImageNet21K, we select the GT label from $S_L$ to compute $f$. For ImageNet-1K ReaL, we randomly sample one label from $S_L$ per image, which we repeat 50 times and report the mean accuracies. We found a maximum standard deviation of $0.1$. For $k$-means, we compute the top 5 predictions by finding the 5 closest centroids.


\section{Experimental results}
\textbf{Implementation details.} Following \citet{ridnik2021imagenet}, we use the preprocessed winter 2021 version of ImageNet21K and its corresponding semantic tree unless otherwise specified. We use the training data to develop the clustering method and evaluate on the validation split as in \cite{scan, temi}. We conduct all the clustering experiments on a single Nvidia A100 GPU with 40GB of VRAM. For TEMI and SCANv2, we use $H$=32 clustering heads and train for 50 and 25 epochs on ImageNet-1K and ImageNet21K, respectively. Different from \cite{temi}, we use a larger batch size (i.e. 4096 on ImageNet-1K instead of 512) for SCANv2 as we observed training instabilities and improved performance. For $k$-means clustering, we use cosine similarity as the distance metric and apply L2 normalization to the cluster centers after each iteration. We report the average clustering top 1 accuracy (ACC) on the validation set from three independent runs. Finally, we report the linear probing accuracy as an upper bound of clustering methods. All hyperparameters and additional clustering metrics are provided in the supplementary material.

\begin{table}
    \centering
\begin{tabular}{l cccccc}
\toprule
 &  \multicolumn{2}{c}{Information}  &   \multicolumn{4}{c}{Accuracy (\%)} \\
 \cmidrule(lr){2-3}   \cmidrule(lr){4-7} 
  & \# samples & \# classes &  \textcolor{lightgray}{Probing} & $k$-means  & TEMI & SCANv2 \\
\midrule
ImageNet-1K  &  1.28M & 1000 &      \textcolor{lightgray}{82.90}         &  62.72  & 66.40 & \textbf{69.79}  \\
ImageNet-1K ODD  &  0.96M & 1000 &  \textcolor{lightgray}{82.05}     & 60.76 & 62.82 & \textbf{65.80}       \\
\hline
\multicolumn{5}{l}{\textit{ImageNet21K subset benchmarks}}  \\
Imbalanced-2K  & 2.55M & 2113 &  \textcolor{lightgray}{67.77}            & 34.78           & \textbf{36.01}   & 34.21  \\
Centered-2K  & 2.34M & 2113 &    \textcolor{lightgray}{68.51}      &    35.22        &  \textbf{36.73} &  34.53 \\


\bottomrule
\end{tabular}
    \caption{Accuracies on imbalanced benchmarks. We use MAE-R  \cite{alkin2024mae_refine} on ImageNet1K benchmarks and iBOT \cite{zhou2021ibot} on ImageNet21K benchmarks. Bold indicates the highest clustering accuracy, and supervised results are marked in \textcolor{lightgray}{grey}.}\label{tab:freq-in21k-splits}
\end{table}

\subsection{Impact of class imbalance}
\textbf{ImageNet-1K.} We first verify the superiority of TEMI and SCANv2 on ImageNet-1K in \Cref{tab:freq-in21k-splits} where we measure at least 5.9\% and 10.0\% relative ACC gain compared to $k$-means using the latest state-of-the-art feature extractor MAE-R \cite{alkin2024mae_refine}. With a well-tuned SCANv2 framework, we achieve a new state-of-the-art clustering ACC on ImageNet-1K of 69.79\%. Notably, this accuracy surpasses a supervised GoogLeNet (2015) \cite{szegedy2015going} as a 68.3\% top-1 ACC has been reported in \cite{iandola2016firecaffe}. We clarify that we increase the mini-batch size from 512 to 4096 to get the state-of-the-art ACC and provide an explanation of why it is necessary in \Cref{sec:discussion}.

\begin{figure}
\begin{center}
  \includegraphics[width=0.99\columnwidth]{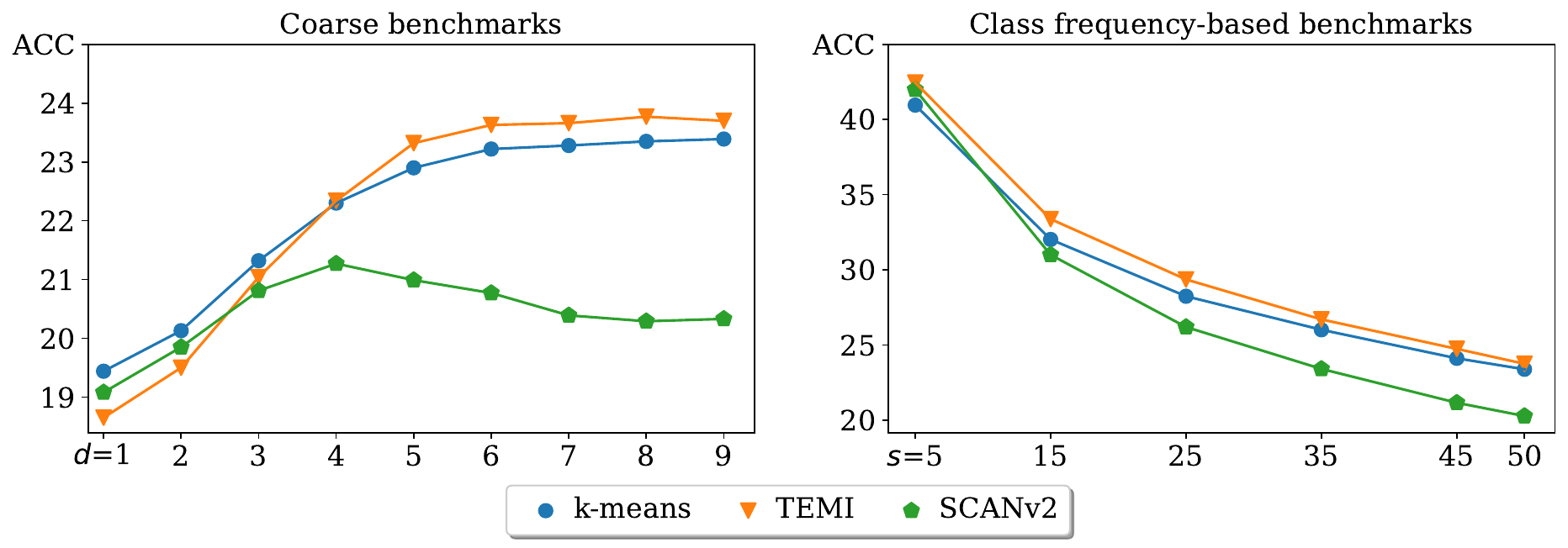}
    \caption{\textbf{Left:} Clustering accuracy in \% (\textit{y-axis}) versus maximum hierarchy depth $d$ (\textit{x-axis}) by mapping each ImageNet21K class to its semantic ancestor at depth $d$. \textbf{Right:} Clustering accuracy in \% (\textit{y-axis}) versus interval $s$ (\textit{x-axis}) centered around the median $m_s =[50-s,50+s]$ based on the ImageNet21K class histogram. Best viewed in color.}
    \label{fig:in21k_benchmarks}
\end{center}
\end{figure}

\textbf{Imbalanced benchmarks.}
For imbalanced datasets such as ImageNet-1K ODD and Imbalanced-2K, the ACC gains between deep clustering methods over $k$-means are reduced. We measure a relative ACC degradation of 3.2\% for $k$-means, 5.4\% for TEMI, and 5.7\% for SCANv2. Similar behavior is observed on the ImageNet21K scale when comparing the Centered-2K versus Imbalanced-2K benchmarks. Even when gradually scaling up the dataset size and the degree of imbalance simultaneously (\Cref{fig:in21k_benchmarks}, right), we observe the same trend: although TEMI consistently outperforms $k$-means, its relative ACC gain diminishes (from a maximum of 4.23\% to 1.6\% using all samples). Based on the above, we state that a) $k$-means is unfairly evaluated solely on balanced clustering benchmarks such as CIFAR and ImageNet-1K, b) $k$-means demonstrates suboptimal performance even on imbalanced benchmarks, and c) the degree of imbalance only partially accounts for the inferior performance of k-means observed in \Cref{tab:freq-in21k-splits} and \Cref{fig:in21k_benchmarks} as well as previous studies \cite{sscn,temi,scan,cpp}. Another veiled aspect is that despite both deep clustering methods enforcing class uniformity, they are applicable beyond the scope of ImageNet-1K, such as large-scale imbalanced datasets.

\begin{table}
    \centering
\begin{tabular}{l cccccc}
\toprule
 &  \multicolumn{2}{c}{Information}  &   \multicolumn{4}{c}{Accuracy (\%)} \\
 \cmidrule(lr){2-3}   \cmidrule(lr){4-7} 
  & \# samples & \# classes & \textcolor{lightgray}{Probing} &  $k$-means  & TEMI  & SCANv2  \\
\midrule
\multicolumn{5}{l}{\textit{ImageNet21K subset benchmarks}}  \\
WordNet leaf   & 7.54M & 7353 &    \textcolor{lightgray}{56.92}           & 27.43  & \textbf{28.08}  & 24.56\\
ImageNet21K leaf & 8.37M  & 8153 & \textcolor{lightgray}{55.31}     & 26.66  &    \textbf{ 27.31} & 23.58\\
\hline
\multicolumn{5}{l}{\textit{Relabeled ImageNet21K benchmarks}}  \\
Coarse $d=4$  & 11.06M  & 8386 &    \textcolor{lightgray}{49.99}      &      22.30 & \textbf{22.34} & 21.27 \\
Leaf HZR  & 11.06M  & 8153 &        \textcolor{lightgray}{52.25}         &   24.48      & \textbf{24.56}   & 22.72  \\
p-HZR  & 11.06M  & 10413 &          \textcolor{lightgray}{49.90}         &   24.27      &  \textbf{24.50} &  21.83 \\
GT labels  & 11.06M & 10450 &       \textcolor{lightgray}{46.72} &  23.38 &  \textbf{23.74} &20.27 \\

\bottomrule
\end{tabular}
    \caption{Accuracies of fine-grained benchmarks based on ImageNet21K. For reference, we report the coarse dataset for a max depth of $d=4$ and the original ImageNet21K (GT).}\label{tab:fine_grained}
\end{table}

\subsection{Impact of class granularity: coarse and fine-grained benchmarks}

In \Cref{fig:in21k_benchmarks} (left), we illustrate the methods' accuracies for different maximum hierarchy depths $d$. Interestingly, $k$-means outperforms TEMI for $d<3$ with a maximum relative difference of 4.1\% for $d=1$. We argue that highly coarse labels are more likely to be captured by isotropic and normally distributed clusters assumed by $k$-means. Conversely, TEMI and SCANv2 rely on fixed NN pairs, which lead to retrieving similar-looking images to the query image \cite{dinov2}, which hinders the coarse clustering performance for superclasses such as ``animal''.

Regarding the considered fine-grained benchmarks in \Cref{tab:fine_grained}, all clustering methods benefit similarly from purging mutually exclusive classes. The highest relative gain is observed in the subsets of WordNet leaf classes, where TEMI and $k$-means improve by 18.29\% and 17.32\% (relative ACC) compared to ImageNet21K. Overall, TEMI outperforms $k$-means on the majority of coarse and fine-grained ImageNet21K-based benchmarks, but the gains are marginal. Based on \Cref{fig:in21k_benchmarks} and \Cref{tab:fine_grained}, we cannot identify any strong ACC discrepancy between clustering methods w.r.t. the class granularity.

\newcommand*{\ppos}{p_\mathsf{pos}}
\newcommand*{\lunif}{\mathcal{L}_\mathsf{uniform}}
\newcommand*{\lalign}{\mathcal{L}_\mathsf{align}}
\newcommand*{\distnpos}{\ppos}
\newcommand*{\pdata}{p_\mathsf{data}}
\newcommand*{\distndata}{\pdata}

\subsection{Results from easy-to-classify benchmarks: model-based ImageNet21K subsets}
All three model-based splits of 1K classes achieve high probing ACC ($>95$\%) using iBOT (pretrained on ImageNet21K) as shown in \Cref{tab:model-splits}. This indicates that the image features of easy-to-classify classes are well-separated, verifying \cite{wang2020understanding}. Intriguingly, we measure a high discrepancy between $k$-means and deep learning methods that is independent of the model used to pick the top-1K classes (CLIP, DINOv2, Supervised). The reported discrepancy does not originate from the degree of imbalance. Compared to TEMI, $k$-means underperforms by a relative accuracy difference of at least 12.9\%. We thus state that well-separated clustering benchmarks may have irregular class shapes \cite{ezugwu2022comprehensive} that cannot always be modeled with $k$-means. Deep clustering methods are more flexible by leveraging the structure of the feature space (through the NN) to capture irregular class shapes. The latter likely leads to a decision boundary of a more arbitrary form compared to $k$-means.

\begin{table}
    \centering
\begin{tabular}{l cccccc}
\toprule
 &  \multicolumn{2}{c}{Information}  &   \multicolumn{4}{c}{Accuracy (\%)} \\
 \cmidrule(lr){2-3}   \cmidrule(lr){4-7}  
 ImageNet21K subsets & \# samples & \# classes & \textcolor{lightgray}{Probing}  &  $k$-means  & TEMI  & SCANv2  \\
\midrule
(1) CLIP-1K  & 1.14M      & 1K    &  \textcolor{lightgray}{95.77}        & 55.05    & \textbf{63.29} & 62.90\\
(2) DINOv2-1K &  1.13M    & 1K    &  \textcolor{lightgray}{95.74}      & 54.24    &   \textbf{61.21} & 61.06   \\
(3) Sup-1K  &  1.12M      & 1K    &  \textcolor{lightgray}{95.96}     & 55.35    &    62.52 & \textbf{62.65} \\
Random 1K   &  1.06M      & 1K    &  \textcolor{lightgray}{78.12}         & 41.50   & \textbf{44.13} & 43.59 \\
ImageNet-1K   & 1.28M     & 1K    &  \textcolor{lightgray}{81.53}         & 41.94 & \textbf{47.32} & 45.76   \\
\hline
Union (1,2,3)  & 1.49M & 1331  & \textcolor{lightgray}{94.62} & 52.14   &  \textbf{59.40} &  57.95  \\
\bottomrule
\end{tabular}
    \caption{Accuracies on model-based ImageNet21K subsets based on different pre-trained models. We use the top-1K classes (highest ACC) of OpenCLIP ViT-G, DINOv2 ViT-g, and a Supervised ViT-L. For the reported accuracies, we use iBOT ViT-L pretrained on ImageNet21K.}
    \label{tab:model-splits}
\end{table}

\subsection{Multi-label clustering evalutions}
As shown in \Cref{tab:metrics}, the top 1 predictions capture relevant additional concepts (coarser labels, coexisting labels, or multiple objects of interest) as measured by top$1\rightarrow$L. We measure a relative top 1$\rightarrow$L ACC increase of more than $13$\% (ImageNet21K) and $8.1$\% (ImageNet1K-ReaL) across clustering methods. In terms of the top 5 ACCs, we find that the non-primary cluster predictions capture meaningful classes, corroborating with \citet{beyer2020done_imagenet}. Unlike probing where single-label supervision is provided during training, clustering methods achieve higher relevant ACC gains in top$5\rightarrow$1 and top$5\rightarrow$L on both benchmarks.


\begin{table}
\centering
\begin{tabular}{l cccc}
\toprule
&   Top$1\rightarrow$1 (\%) & Top$1\rightarrow$L (\%) & Top$5\rightarrow$1 (\%) & Top$5\rightarrow$L (\%)  \\
\hline
\multicolumn{5}{l}{\textit{ImageNet21K using iBOT ViT-L}}  \\
$k$-means  & 23.4 & 26.6 (13.5\%) & 39.9 (70.7\%) & 46.2 (97\%)    \\
TEMI    & 23.7 & 27.1 (\textbf{13.8}\%) & 42.7 (\textbf{79.5}\%) & 48.8 \textbf{(105\%)}   \\ 
SCANv2    & 20.3 & 22.9 (13.0\%) & 31.8 (56.8\%) & 37.8 (86\%)     \\
\textcolor{lightgray}{Linear probing}    & \textcolor{lightgray}{47.3} & \textcolor{lightgray}{51.3 (8.3 \%)} & \textcolor{lightgray}{76.4 (61.3\%)} & \textcolor{lightgray}{79.8 (68\%)}    \\
\hline
\multicolumn{5}{l}{\textit{ImageNet1K-ReaL using MAE-R ViT-H}}  \\
$k$-means   &      62.6 & 68.2 (\textbf{8.9}\%) & 78.2 (\textbf{24.9}\%) & 83.1 (\textbf{32.6}\%) \\
TEMI        &      67.3 & 73.1 (8.5\%) & 82.0 (21.8\%) & 86.8 (28.8\%)  \\
SCANv2        &      70.2 & 75.9 (8.1\%) & 84.6 (20.5\%) & 89.4 (27.3\%)  \\
\textcolor{lightgray}{Linear probing}         &      \textcolor{lightgray}{81.1} & \textcolor{lightgray}{88.2 (8.8\%)} & \textcolor{lightgray}{94.4 (16.4\%)} & \textcolor{lightgray}{97.3 (20.0\%)}  \\
\bottomrule 
\end{tabular}
\caption{Multi-label clustering accuracies evaluated using the semantic ancestors on ImageNet21K and ImageNet-1K ReaL \cite{beyer2020done_imagenet}. The parentheses refer to the relative improvement over the top 1$\rightarrow$1. We report mean accuracies on ImageNet-1K ReaL after 50 random samples from the label set $S_L$.}
\label{tab:metrics}
\end{table}




\section{Discussion, limitations and future work} \label{sec:discussion}

\textbf{Class separability and clustering benchmarks.} To further investigate the reported discrepancy between $k$-means and deep clustering methods (\Cref{tab:model-splits}), we measure the GT alignment score \cite{wang2020understanding}, the silhouette score \cite{rousseeuw1987silhouettes} and the Davies-Bouldin score (DBS) \cite{dbs1979cluster} using the iBOT feature extractor. All metrics highly correlate with the probing ACC, specifically $R^2>0.97$ (see Supp.), and point to the fact that model-based subsets are the most well-separated benchmarks. We highlight that model-based filtering is an automatic way to create new benchmarks without any explicit constraint on the imbalance or granularity level. Still, we assume that a feature extractor trained on external data (e.g. CLIP) is available. 


\textbf{Are we done with ImageNet-1K?} No, but additional large-scale clustering benchmarks should be taken into account. Limitations and idiosyncrasies of ImageNet-1K have been pointed out in a series of classification-centered studies \cite{beyer2020done_imagenet, wen2022misclassified_ImageNet}. From the perspective of clustering, ImageNet-1K is substantially closer to a random split of 1K ImageNet21K classes in terms of linear separability (\Cref{tab:model-splits}). Given that ImageNet-1K is balanced (\Cref{tab:freq-in21k-splits}), a non-inherent limitation of clustering, we state it should not be used in isolation as the core large-scale benchmark.

\textbf{Sensitivity of SCANv2 to the mini-batch size.} In contrast to \citet{temi}, we identified that SCANv2 outperforms TEMI on ImageNet-1K (\Cref{tab:freq-in21k-splits}). Yet SCANv2 performs consistently inferior on the large-scale benchmarks (\Cref{tab:fine_grained}), which is partially attributed to its sensitivity to the mini-batch size $B$. In particular, the entropy term in \Cref{eq:scan-loss} requires $B$ to be sufficiently larger than the number of clusters $C$ to approximate the cluster distribution $p(c)$. The latter requires a lot of extra memory as the space complexity scales linearly $O(C \times H)$ w.r.t. the number of clusters $C$ and heads $H$. This computational limitation becomes quite significant on large-scale benchmarks as shown in \Cref{tab:fine_grained}. 

\begin{figure}
\begin{center}\includegraphics[width=0.99\columnwidth]{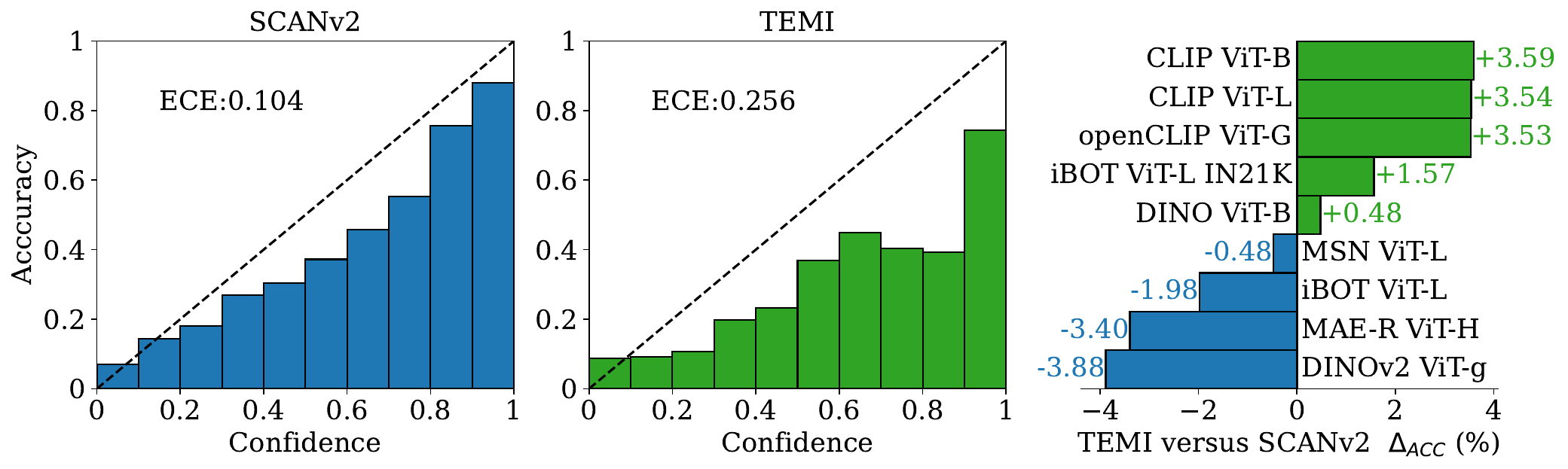}
    \caption{\textbf{Left and middle:} Calibration plots and expected calibration error (ECE $\downarrow$) for SCANv2 and TEMI on ImageNet-1K using MAE-R ViT-H \cite{alkin2024mae_refine}. \textbf{Right:} Clustering ACC difference between TEMI and SCANv2 on ImageNet-1K across various pre-trained feature extractors. IN21K refers to ImageNet21K, and CLIP ViT-B and ViT-L use the weights from \cite{radford2021clip} .} 
    \label{fig:backbones}
\end{center}
\end{figure}

\textbf{Calibration analysis of deep clustering methods.} Unlike $k$-means, deep clustering methods provide a measure of confidence as measured by the maximum softmax probability (MSP) \cite{hendrycks2016msp}. In \Cref{fig:backbones} (left and middle), we show that even though TEMI yields more discriminative predictions (91.9\% versus 80.2\% mean MSP), the predictions of SCANv2 are better calibrated on ImageNet-1K. Again, this is attributed to the entropy regularization in \Cref{eq:scan-loss}, which explicitly down-weights images with high confidence in the mini-batch. We believe that the confidence of well-calibrated clustering approaches can be leveraged in many applications, such as dataset pruning and semi-automatic data annotation.

\textbf{Dependence on feature extractors.} A limitation of this study is the dependence of feature-based clustering methods on the learned representations. Both TEMI and SCANv2 consistently outperform $k$-means across feature extractors on ImageNet-1K. However, as shown in \Cref{fig:backbones} (right), no method consistently surpasses when varying pre-trained feature extractors. Based on that, we believe there is room for further investigation in feature-based clustering.  

\section{Conclusion}
In this paper, a comprehensive experimental study of feature-based clustering approaches and large-scale benchmarks was conducted. New clustering benchmarks based on ImageNet21K were created, focusing on the effects of individual factors, such as class imbalance, class granularity, easily separable classes, and the ability to capture multiple labels. TEMI and SCANv2 were shown to outperform $k$-means on the majority of introduced benchmarks. However, we found marginal differences on the largest scale benchmarks. An inferior clustering performance was observed by $k$-means on easy-to-classify benchmarks. It was demonstrated that clustering methods can capture multiple GT labels, often with meaningful secondary predictions such as coarser classes. We believe that the new set of benchmarks will help future clustering approaches disentangle individual factors on large scales.

\bibliographystyle{plainnat} 
\bibliography{refs}

\appendix



\section{Licences, links, and availability of datasets}
Below are the datasets, licenses, codebases, and relevant information to enforce transparency and facilitate reproducibility. The code and all benchmarks are submitted in the review as supplementary material and will be made publicly available upon acceptance of the manuscript.
\begin{itemize}
    \item We use the ImageNet21K train and validation split of \citet{ridnik2021imagenet} on the preprocessed winter 2021 version of ImageNet\cite{deng2009imagenet,russakovsky2015imagenet} as well as the corresponding semantic tree, distributed under an MIT license. More information can be found at \url{https://github.com/Alibaba-MIIL/ImageNet21K}. \citet{ridnik2021imagenet} have enabled direct downloading of ImageNet-21K-P via the official ImageNet site \url{https://image-net.org/request}.
    \item The WordNet hierarchy \cite{miller1995wordnet} was loaded from the NLTK library \cite{xue2011nltk}, available at \url{https://github.com/nltk/nltk}, under Apache License Version 2.0.
    \item ImageNet-1K \cite{deng2009imagenet,russakovsky2015imagenet} is publicly available and can be download from \url{https://www.kaggle.com/competitions/imagenet-object-localization-challenge/data}.
    \item \citet{temi} have published their code for TEMI and the SCANv2 baselines considered in this work at \url{https://github.com/HHU-MMBS/TEMI-official-BMVC2023}, distributed under under Apache License Version 2.0.
    \item \citet{beyer2020done_imagenet} provide the reassessed labels of ImageNet-1K validation set in the following link \url{https://github.com/google-research/reassessed-imagenet}.
    \item We have used the openCLIP models \cite{openclip,cherti2023openclip}, publicly available at \url{https://github.com/mlfoundations/open_clip}  (MIT License), as well as the OpenAI CLIP weights available \cite{radford2021clip} at \url{https://github.com/openai/CLIP} (MIT License).
    \item We use the distributed version of $k$-means available in the faiss library \cite{douze2024faiss} \url{https://github.com/facebookresearch/faiss}.
\end{itemize}

\section{Additional discussion points}
\subsection{Linear separability metrics.} 
\begin{figure}[h]
\begin{center}\includegraphics[width=1\columnwidth]{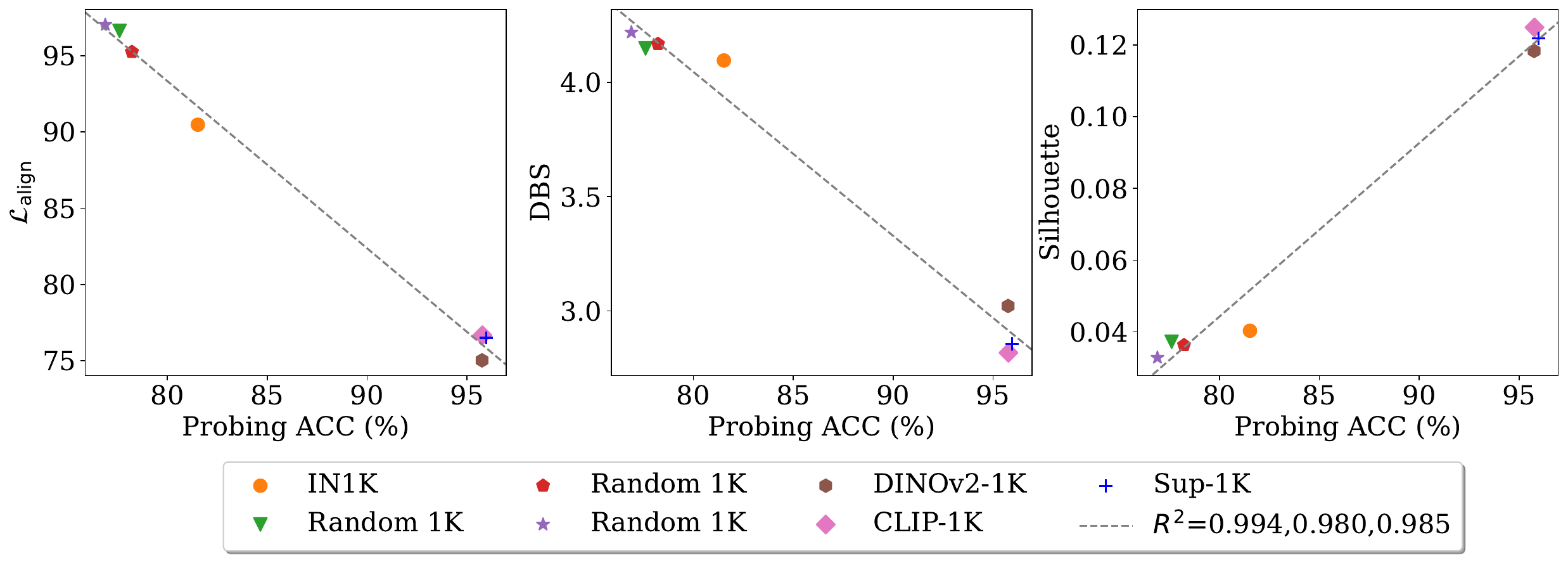}
    \caption{We measure the dependence of the linear probing accuracy \textit{x-axis} versus the alignment score $\mathcal{L}_\mathsf{align}$ (\textbf{left}), the Davies-Bouldin score (DBS) \cite{dbs1979cluster} (\textbf{middle}) and the Sihlouette score (\textbf{right}) using iBOT ViT-L trained on ImageNet21K. $R^2$ is the coefficient of determination. Best viewed in color.} 
    \label{fig:align}
\end{center}
\end{figure}

In \Cref{fig:align}, we observe that all metrics (linear probing ACC, $\mathcal{L}_\mathsf{align}$, Davies-Bouldin score, Silhouette) of class separability are strongly correlated ($R^2>0.98$). First, the model-based ImageNet-21K benchmarks are well-clustered in the feature space of iBOT. Second, ImageNet-1K is much closer to a random split of 1K class from ImageNet21K. $\mathcal{L}_\mathsf{align}$ is the alignment of features that share the GT same label ($\distnpos$) as $\mathcal{L}_\mathsf{align}(g)=\mathbb{E}_{(x_1, x_2) \sim \distnpos}{\norm{g(x_1) - g(x_2)}_2^2}$ \cite{wang2020understanding}. \citet{wang2020understanding} have used augmentations of the same image as positive pairs ($\distnpos$) while we use true positive nearest neighbors, meaning images that share the same labels.

\subsection{Label reassessment and p-HZR.} An increase in linear separability w.r.t the refined labels compared to the GT suggests that refined labels align better with the learned features. Surprisingly, p-HZR yields a relative probing ACC gain of 6.8\% while keeping 99.6\% of the classes. A similar probing ACC was measured for the coarse benchmark for $d=4$ (80.2\% of the classes). We identify many semantic links in WordNet that do not align with sensible visual-based concepts and likely limit the effectiveness of such approaches. Examples include person$\rightarrow$father-in-law, animal$\rightarrow$adult. We hypothesize that a more visually aligned tree would improve automatic tree-based label reassessment. Nonetheless, deriving a single gold standard label \cite{hollink2019fruit,beyer2020done_imagenet} (the granularity level at which humans assign the GT label quicker and with greater accuracy) requires humans in the loop. That is why we find multi-label clustering evaluations more informative on large-scale datasets, which has not been sufficiently investigated by previous works.

\subsection{Increasing the number of TEMI nearest neighbors for the coarse ImageNet21K benchmarks.} 
\begin{figure}[h]
\begin{center}
\includegraphics[width=0.9\columnwidth]{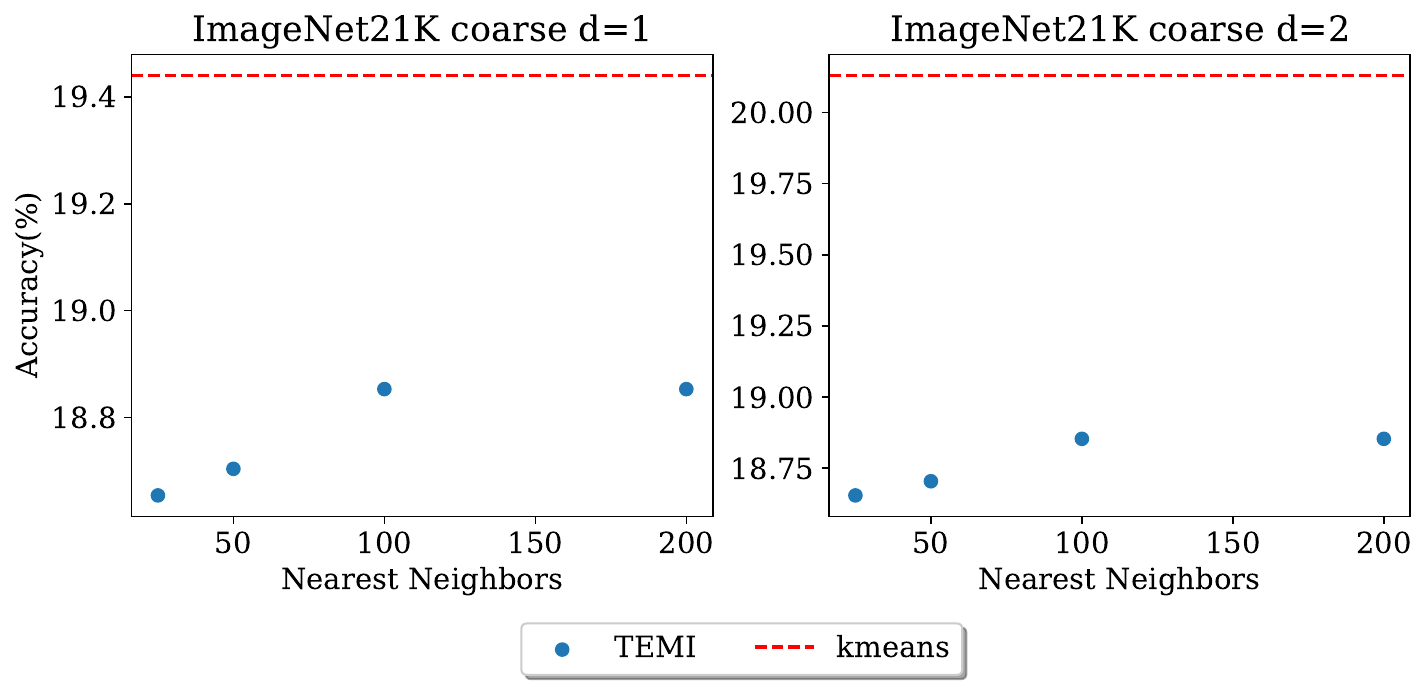}
    \caption{ We show that increasing the number of NN increases the ACC of TEMI but $k$-means still outperforms TEMI on the coarse ImageNet21K dataset benchmarks. \textbf{Left:} ImageNet21K with a maximum class depth of 1 (root labels on the semantic tree). \textbf{Right:} ImageNet coarse benchmark with maximum class depth of 2.} 
    \label{fig:coarse}
\end{center}
\end{figure}

As discussed in the main paper, we observe a slight increase in the performance of TEMI when increasing the number of nearest neighbors to capture the superclass groups created from the coarse annotations of the semantic tree. However, the performance increase in TEMI accuracy is marginal, and $k$-means outperforms TEMI when the class labels are highly coarse. Our experimental results are depicted in \Cref{fig:coarse}.

\subsection{Overclustering results of deep clustering methods on ImageNet-1K} 
\begin{figure}
\begin{center}\includegraphics[width=0.8\columnwidth]{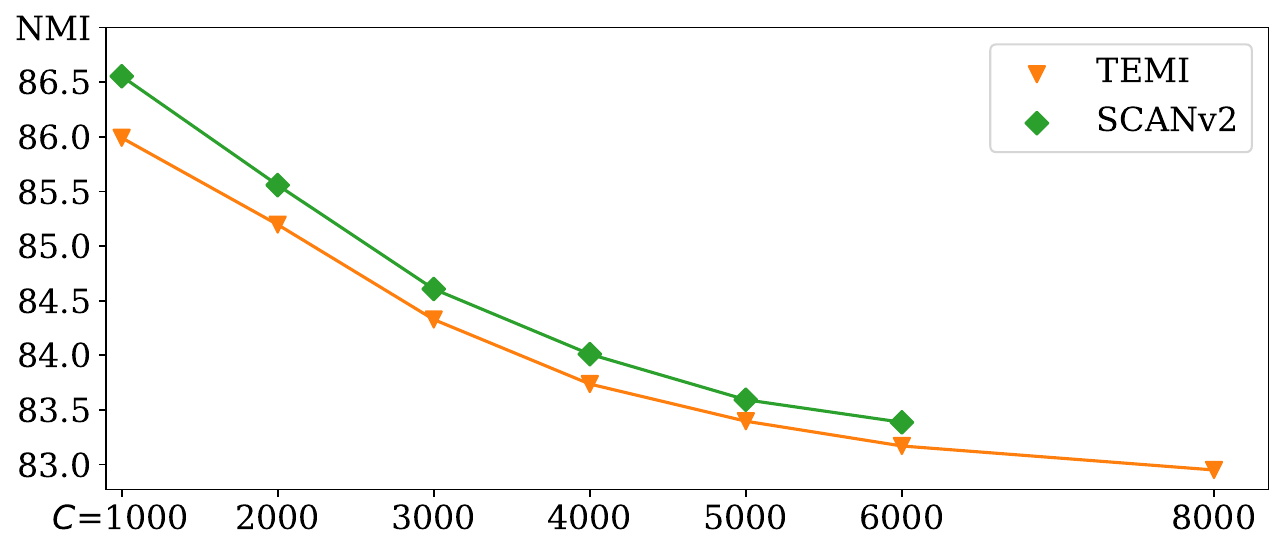}
    \caption{Normalized mutual information (NMI $\uparrow$, \textit{y-axis}) versus the number of clusters $C$ (\textit{x-axis}) on ImageNet-1K. SCANv2 was out of memory in the considered hardware of 40GB VRAM for $C>6000$.} 
    \label{fig:overcluster}
\end{center}
\end{figure}

We measure the effect of the prior knowledge of the number of clusters $C$ in \Cref{fig:overcluster}. The absolute performance difference between TEMI and SCANv2 progressively diminishes for large $C$. Additionally, we find that SCANv2 is out of memory since a larger batch size (4096 versus 512 for TEMI) is required to get the optimal performance and avoid training instabilities. 

\subsection{CLIP zero-shot classification and calibration analysis on ImageNet21K}
\begin{figure}[h]
\begin{center}
  \includegraphics[width=1\columnwidth]{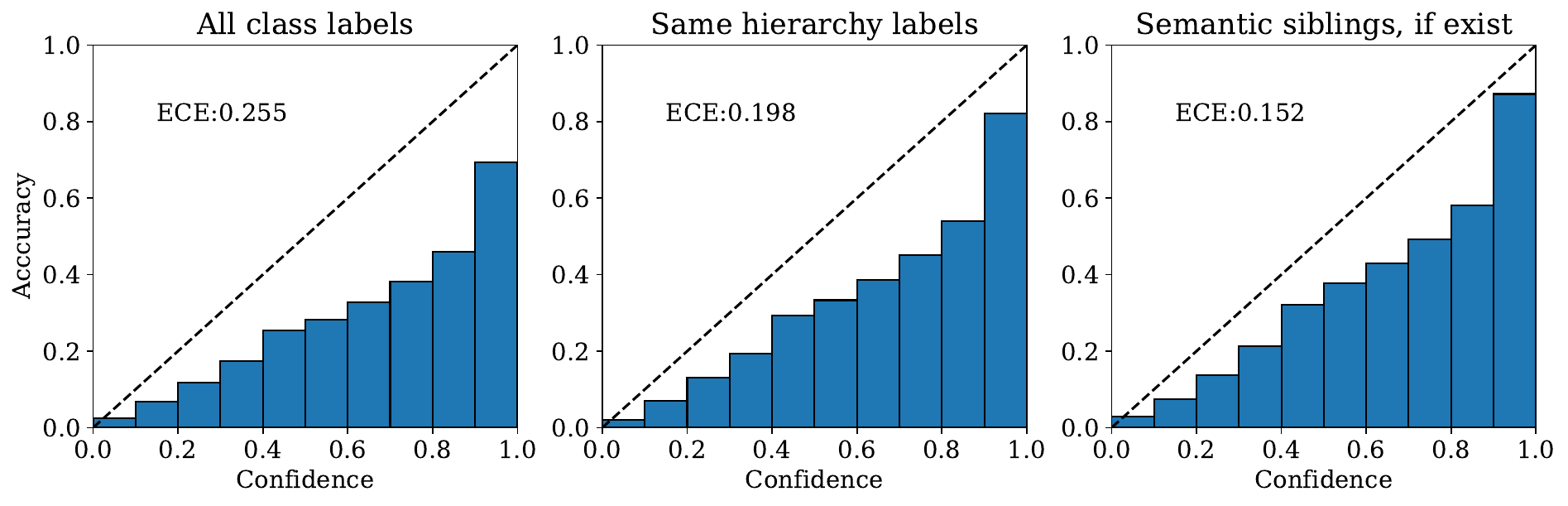}
    \caption{\textbf{Left:} Zero-shot classification using OpenCLIP with all 10450 class names on ImageNet21K (36.02\% mean accuracy). \textbf{Middle:} We restrict the candidate labels for zero-shot classification to the labels of the same hierarchy (55.10\% mean accuracy). \textbf{Right:} We further restrict the candidate labels to the semantic siblings based on the semantic tree (68.35\% mean accuracy). When there are no semantic siblings, we use the classes from the same hierarchy. ECE stands for expected calibration error.} 
    \label{fig:calibration}
\end{center}
\end{figure}

We perform a calibration analysis \cite{guo2017calibration} on the zero-shot classification of openCLIP ViT-G on ImageNet21K) in \Cref{fig:calibration}. We find that the baseline zero-shot ACC using all 10450 classes of 36.02\% can be increased to 55.10\% when considering the subset of class labels from the same hierarchy. When further restricting the candidate labels to the semantic siblings of the GT class, if available, we measure an accuracy of 68.35\%. Siblings refer to all class labels that share the same semantic ancestor as the GT. When semantic siblings are not available, we use the labels from the same hierarchy. In parallel, we find that CLIP is better calibrated when restricting the number of candidate labels to siblings, as measured by a relative decrease of 40\% in the expected calibration error (ECE) \cite{guo2017calibration}. We measure a mean confidence (maximum softmax probability \cite{hendrycks2016msp} on the image-text similarities) of 83.44\% versus 61.4\% (baseline) using the semantic siblings. 

\subsection{Confidence histogram of deep clustering methods on ImageNet-1K}
\begin{figure}[h]
\begin{center}
  \includegraphics[width=1\columnwidth]{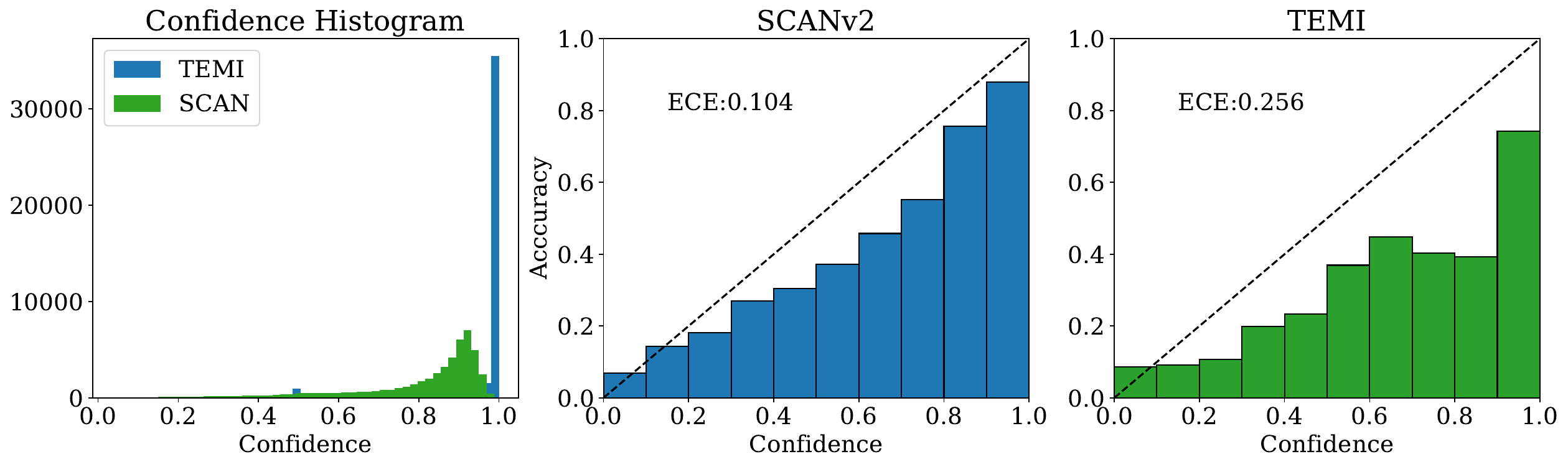}
    \caption{\textbf{Left:} Histogram of maximum softmax probabilities (confidences) on ImageNet-1K. \textbf{Middle, right:} Calibration plots for SCANv2 and TEMI. ECE stands for expected calibration error.} 
    \label{fig:histogram_calibration}
\end{center}
\end{figure}

While we have shown in the main text that SCANv2 is better calibrated than TEMI, their confidence histogram looks significantly different on ImageNet-1K, as illustrated in \Cref{fig:histogram_calibration}. TEMI produces more discriminative predictions likely due to the pointwise mutual information objective. However, this makes its predictions less calibrated. We leave this finding for future investigation.

\section{Additional implementation details}

\subsection{Additional information regarding parent hierarchical zero-shot label refining (p-HZR) on ImageNet21K.}
The proposed method for reannotating ImageNet21K (p-HZR) only reduces the number of classes by 37 (train and validation), which requires excluding only 484 training samples ($<0.01$\% of the samples). Overall, we are able to refine $\approx42$\% of the training GT labels that are not leaf classes, which translates to more than 10\% of ImageNet21K.

\subsection{Zero-shot classification.} During all our zero-shot experiments with CLIP, we first match the WordNet's unique identifier to one of the words in its set of synonyms, typically called lemmas, such as n00007846$\rightarrow$[person, individual, someone, soul]. To this end, we perform zero-shot classification across lemmas for all images of the same class and subsequently apply majority voting. This step effectively reflects the linguistic biases of the pre-training data used to train the vision-language model, namely LAION-2B \cite{schuhmann2022laionb}. For example, CLIP prefers ``artefact" instead of ``artifact". We use a subset of 7 out of the 80 prompts initially proposed in \cite{radford2021clip}, namely "itap of a \{label\}", "a bad photo of the \{label\}", "an origami \{label\}", "a photo of the large \{label\}", "a \{label\} in a \{label\} video game", "art of the \{label\}", "a photo of the small \{label\}", based on subsequent prompt engineering analysis of the same authors\footnote{\url{https://github.com/openai/CLIP/blob/main/notebooks/Prompt_Engineering_for_ImageNet.ipynb}}.

\subsection{Linear probing.} For the linear probing experiments, we perform a grid search on a set of learning rates and weight decays. We train the linear layer with stochastic gradient descent for 100 epochs without augmentations and report the highest validation accuracy value obtained. We use the values of \{$10^{-5}, 5 \times 10^{-5}, 10^{-4}, 5\times10^{-4}, 10^{-3}, 5\times10^{-3}$\} and \{$0, 10^{-3}, 10^{-5}$\} for the learning rate and weight decay, respectively.

\subsection{Dataset information for the considered ImageNet21K subsets}
\Cref{tab:sup-extra-tables-dataset} depicts all dataset-related information for the considered frequency-based benchmarks and coarse datasets. We highlight that all datasets have been created based on the preprocessed ImageNet21K winter 2021 version from \citet{ridnik2021imagenet}. 

\begin{table}
    \centering
\begin{tabular}{l cccccc}
\toprule
 &  \multicolumn{2}{c}{Information}  &   \multicolumn{4}{c}{Accuracy (\%)} \\
 \cmidrule(lr){2-3}   \cmidrule(lr){4-7} 
  & \# samples & \# classes &  \textcolor{lightgray}{Probing} & $k$-means  & TEMI & SCANv2 \\
\midrule
\multicolumn{5}{l}{\textit{ImageNet21K frequency-based benchmarks}}  \\
Tiny ($m_{s}=[45, 55]$)      & 1.17M & 1060  &     \textcolor{lightgray}{76.98}            & 40.96           & \textbf{42.46}   & 41.99  \\
Small    ($m_{s}=[35, 65]$)    & 3.46M & 3146 &    \textcolor{lightgray}{63.71}            & 32.02           & \textbf{33.38}   & 30.99  \\
 Base ($m_{s}=[25, 75]$)         & 5.67M & 5230 &  \textcolor{lightgray}{56.53}            & 28.24           & \textbf{29.36}   & 26.19  \\
Large   ($m_{s}=[15, 85]$)      & 7.79M & 7326 &   \textcolor{lightgray}{52.04}            & 26.01           & \textbf{26.70}   & 23.40  \\
Huge    ($m_{s}=[5, 95]$)    & 9.88M & 9405 &      \textcolor{lightgray}{48.57}            & 24.11           & \textbf{24.74}   & 21.15  \\
\hline
\hline
\multicolumn{5}{l}{\textit{Coarse ImageNet21K benchmarks}}  \\
Coarse $d=1$ & 11.06M & 3987  & \textcolor{lightgray}{51.56} & \textbf{19.44} & 18.65 & 19.08 \\
Coarse $d=2$ & 11.06M & 5086  & \textcolor{lightgray}{51.97} & \textbf{20.13} & 19.50 & 19.85 \\
Coarse $d=3$ & 11.06M & 6772  & \textcolor{lightgray}{51.48} & \textbf{21.32} & 21.04 & 20.81 \\
Coarse $d=4$ & 11.06M & 8386  & \textcolor{lightgray}{49.99} & 22.30 & \textbf{22.34} & 21.27 \\
Coarse $d=5$ & 11.06M & 9429  & \textcolor{lightgray}{48.74} & 22.90 & \textbf{23.32} & 20.99 \\
Coarse $d=6$ & 11.06M & 9997  & \textcolor{lightgray}{47.78} & 23.22 & \textbf{23.63} & 20.77 \\
Coarse $d=7$ & 11.06M & 10241 & \textcolor{lightgray}{47.19} & 23.28 & \textbf{23.66} & 20.39 \\
Coarse $d=8$ & 11.06M & 10351 & \textcolor{lightgray}{46.91} & 23.35 & \textbf{23.77} & 20.29 \\
\hline
ImageNet21K   & 11.06M & 10450 & \textcolor{lightgray}{46.72}          & 23.38              & \textbf{23.74} & 20.27 \\

\bottomrule
\end{tabular}
    \caption{We use MAE-R  \cite{alkin2024mae_refine} on ImageNet1K benchmarks and iBOT \cite{zhou2021ibot} on ImageNet21K benchmarks. Bold indicates the highest clustering accuracy, and supervised results are marked in \textcolor{lightgray}{grey}.}\label{tab:sup-extra-tables-dataset}
\end{table}

\section{More reassessed ImageNet21K samples from p-HZR}
Below, we have collected some reassessed ImageNet21K samples using p-HZR to show the effectiveness of not always refining the GT label to a more fine-grained level.

\begin{figure}[b]
\begin{center}
\includegraphics[width=1\columnwidth]{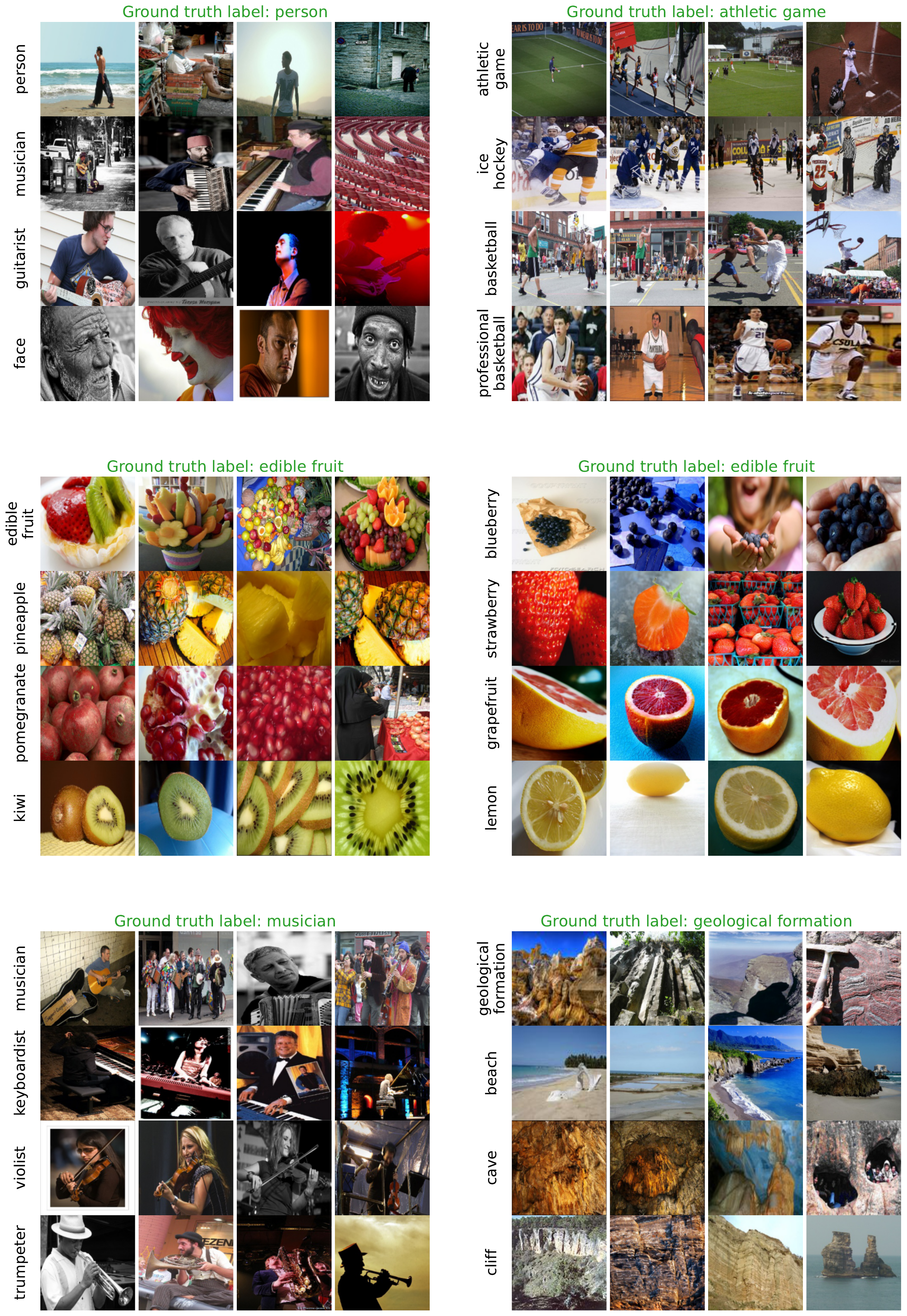}
    \caption{Reassessed ImageNet21K samples using p-HZR. Each subplot's title shows the originally annotated ground truth label. Rows display the reassessed labels. Samples were not reannotated if the reassessed label matched the ground truth label.} 
    \label{fig:hzr1}
\end{center}
\end{figure}
\begin{figure}[b]
\begin{center}
\includegraphics[width=1\columnwidth]{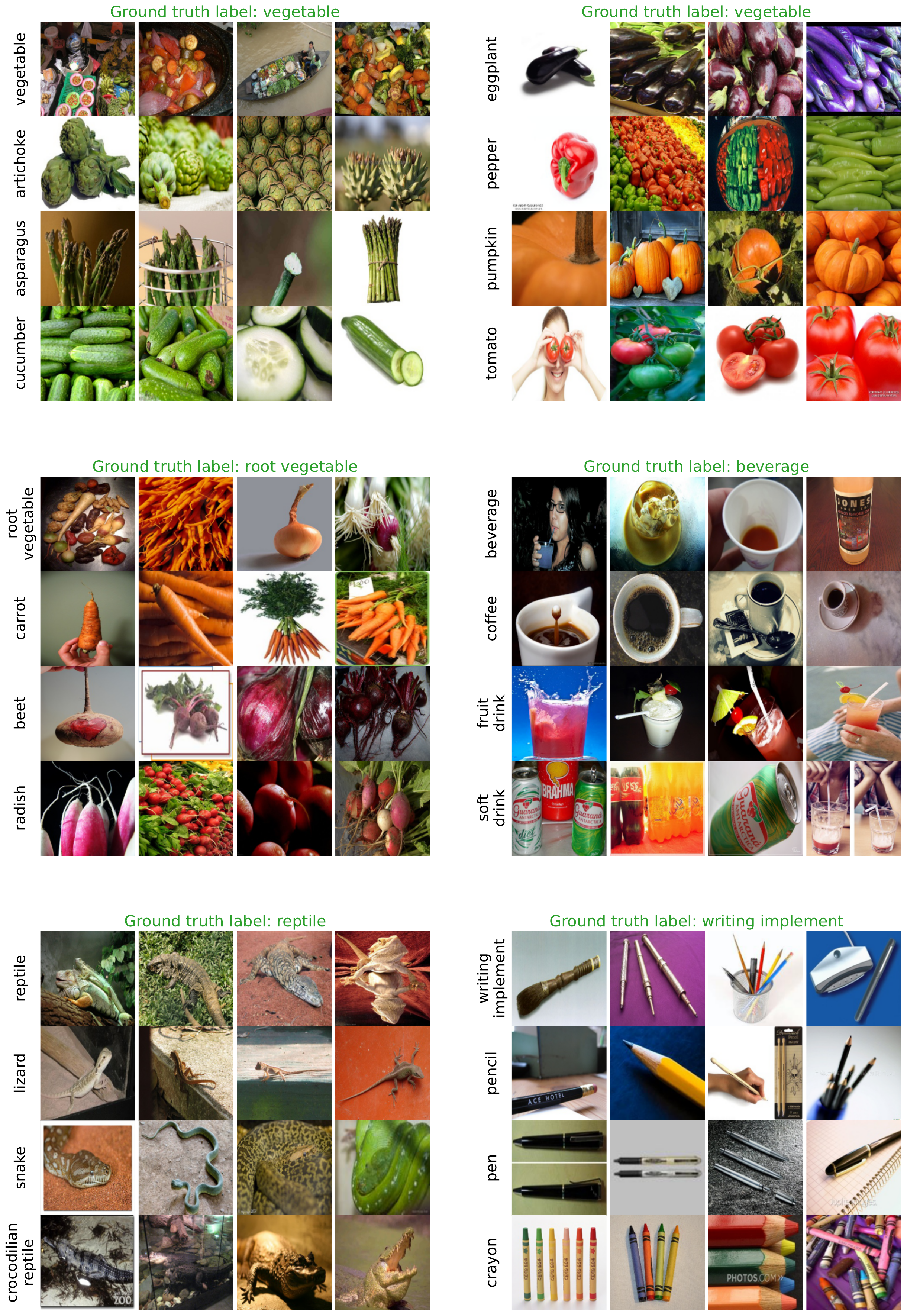}
    \caption{Reassessed ImageNet21K samples using p-HZR. Each subplot's title shows the originally annotated ground truth label. Rows display the reassessed labels. Samples were not reannotated if the reassessed label matched the ground truth label.} 
    \label{fig:hzr1}
\end{center}
\end{figure}
\begin{figure}[b]
\begin{center}
\includegraphics[width=1\columnwidth]{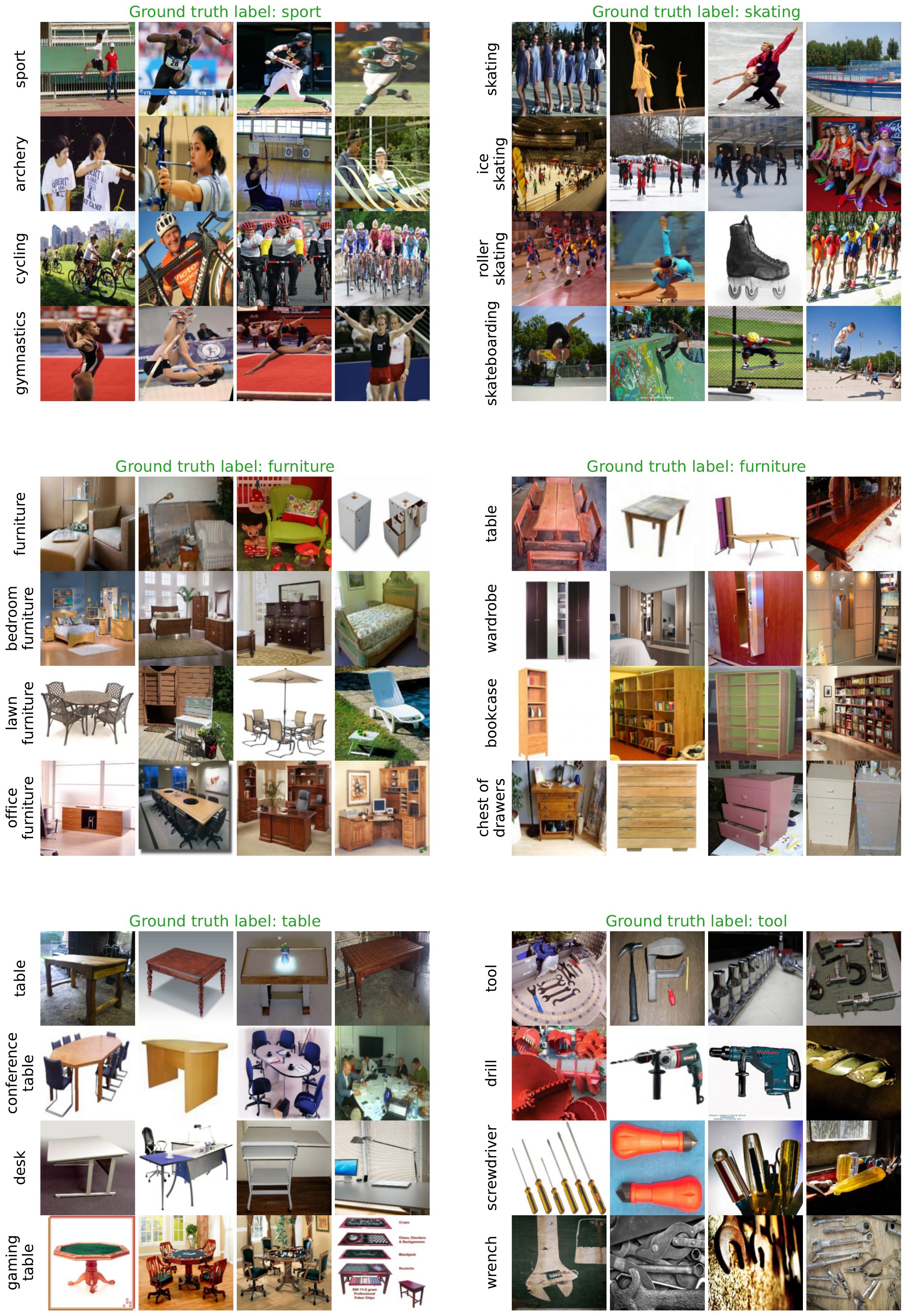}
    \caption{Reassessed ImageNet21K samples using p-HZR. Each subplot's title shows the originally annotated ground truth label. Rows display the reassessed labels. If the reassessed label matches the ground truth label, the samples were not reassessed.} 
    \label{fig:hzr1}
\end{center}
\end{figure}



\end{document}